%% file: root.tex
\documentclass[letterpaper, 10 pt, conference]{ieeeconf}  

\IEEEoverridecommandlockouts                              

\title{\LARGE \bf
Learning Neuro-Symbolic Relational Transition Models\\for Bilevel Planning
}

\author{\textbf{Rohan Chitnis$^*$, Tom Silver$^*$, Joshua B. Tenenbaum, Tom\'as Lozano-P\'erez, Leslie Pack Kaelbling}\\MIT Computer Science and Artificial Intelligence Laboratory\\\texttt{\{ronuchit, tslvr, jbt, tlp, lpk\}@mit.edu}
\thanks{$^*$First two authors contributed equally.}
}

\input{preamble.tex}

\input{variables.tex}

\begin{document}

\bibliographystyle{IEEEtran}

\maketitle
\thispagestyle{empty}
\pagestyle{empty}

\begin{abstract}
\input{abstract.tex}
\end{abstract}

\input{introduction.tex}
\input{related.tex}
\input{problem.tex}

\input{representation.tex}

\input{planning.tex}
\input{learning.tex}
\input{experiments.tex}
\input{conclusion.tex}


\bibliography{references}

\end{document}

%% file: preamble.tex
\usepackage{color}
\usepackage{amssymb}
\usepackage{amsmath}
\usepackage{tikz}
\usepackage{chngcntr}
\usepackage[makeroom]{cancel}
\usetikzlibrary{arrows,decorations.markings}

\usepackage{booktabs}
\usepackage{graphicx}
\usepackage{mathtools}
\usepackage{bbm}

\usepackage{caption}

\usepackage{algorithm2e}
\let\oldnl\nl
\newcommand{\nonl}{\renewcommand{\nl}{\let\nl\oldnl}}
\makeatletter
\newcommand{\removelatexerror}{\let\@latex@error\@gobble}
\makeatother

\newenvironment{tightlist}%
{\begin{list}{$\bullet$}{%
    \setlength{\topsep}{0in}
    \setlength{\partopsep}{0in}
    \setlength{\itemsep}{0in}
    \setlength{\parsep}{0in}
    \setlength{\leftmargin}{1.5em}
    \setlength{\rightmargin}{0in}
}
}%
{\end{list}
}

\newcommand{\secref}[1]{Section \ref{#1}}

\newcommand{\figref}[1]{Figure~\ref{#1}}
\newcommand{\algref}[1]{Algorithm~\ref{#1}}
\newcommand{\algrefshort}[1]{Alg.~\ref{#1}}
\newcommand{\tabref}[1]{Table~\ref{#1}}

\newtheorem{defn}{Definition}

\counterwithin*{thm}{section}

\usepackage{mdframed}
\newmdtheoremenv[nobreak=true]{def2}{Definition}

\def\thickhline{%
  \noalign{\ifnum0=`}\fi\hrule \@height \thickarrayrulewidth \futurelet
  \reserved@a\@xthickhline}
\def\@xthickhline{\ifx\reserved@a\thickhline
              \vskip\doublerulesep
              \vskip-\thickarrayrulewidth
             \fi
      \ifnum0=`{\fi}}

\newlength{\thickarrayrulewidth}

%% file: variables.tex
\newcommand{\fail}{\text{fail}}
\newcommand{\ofail}{\O_\fail}

\renewcommand{\P}{\mathcal{P}}
\renewcommand{\O}{\mathcal{O}}

\renewcommand{\S}{\mathcal{S}} 
\newcommand{\A}{\mathcal{A}}
\newcommand{\T}{\mathcal{T}}

\newcommand{\N}{\mathcal{N}}
\newcommand{\D}{\mathcal{D}}

\DeclarePairedDelimiterX{\infdivx}[2]{(}{)}{%
  #1\;\delimsize\|\;#2%
}

\newcommand{\param}{O}

%% file: abstract.tex
In robotic domains, learning and planning are complicated by continuous state spaces, continuous action spaces, and long task horizons. In this work, we address these challenges with Neuro-Symbolic Relational Transition Models (NSRTs), a novel class of models that are data-efficient to learn, compatible with powerful robotic planning methods, and generalizable over objects. NSRTs have both symbolic and neural components, enabling a bilevel planning scheme where symbolic AI planning in an outer loop guides continuous planning with neural models in an inner loop. Experiments in four robotic planning domains show that NSRTs can be learned very data-efficiently, and then used for fast planning in new tasks that require up to 60 actions and involve many more objects than were seen during training.

%% file: introduction.tex
\section{Introduction}
\label{sec:intro}

\begin{figure*}[t]
  \scriptsize
  \centering
  \noindent
    \includegraphics[width=\textwidth]{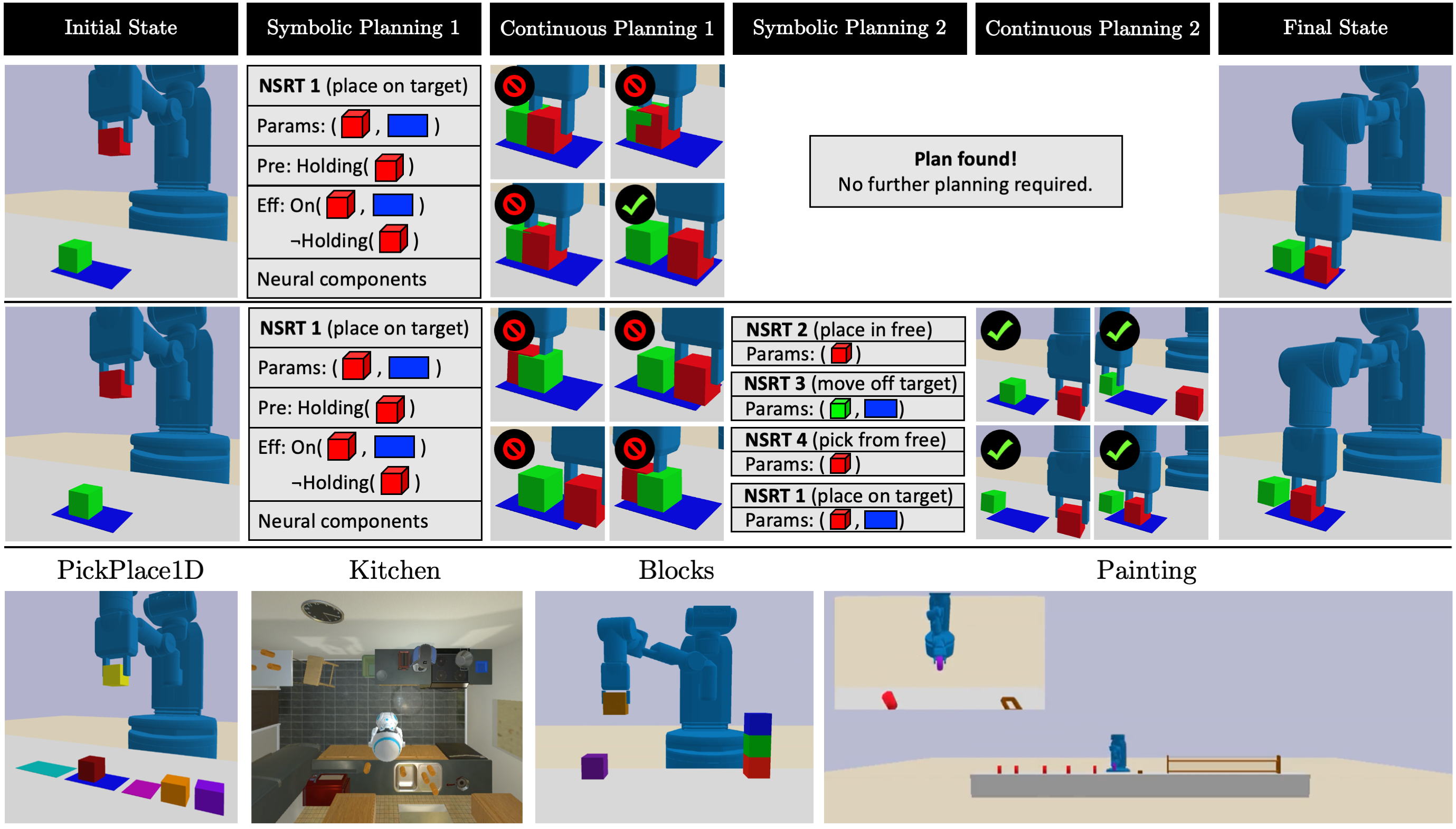}
    \caption{\small We propose Neuro-Symbolic Relational Transition Models (NSRTs). (Top row) Given the goal of placing the red block completely into the blue target region, we first perform AI planning with the symbolic NSRT components to find a one-step symbolic plan. The Continuous Planning 1 column shows various ways in which the agent attempts to \emph{refine} this one-step symbolic plan into a ground action, using the neural components of (ground) NSRT 1; it finds a collision-free refinement, shown in the Final State column. (Middle row) Here, the green block is initially in a slightly different position, so the red block has no room to be placed into the blue target region. The initial symbolic plan is the same. However, this symbolic plan is not \emph{downward refinable}, so Continuous Planning 1 fails. The agent then continues on to consider a four-step symbolic plan that first moves the green object away (Symbolic Planning 2 column), which is successfully refined in the Continuous Planning 2 column. This example illustrates that in the presence of complex geometric constraints which make symbolic abstractions lossy, integrated symbolic and continuous reasoning is necessary. (Bottom row) Screenshots of our four robotic planning environments. Kitchen uses the AI2-THOR simulator~\cite{ai2thor}; the others use PyBullet~\cite{pybullet}.}
  \label{fig:teaser}
\end{figure*}

For robots to plan effectively, they will need to contend with continuous state spaces, continuous action spaces, and long task horizons (\figref{fig:teaser}, bottom row).
Symbolic AI planning techniques are able to solve tasks with very long horizons, but typically assume discrete, factored spaces~\cite{helmert2006fast}.
Neural network-based approaches have shown promise in continuous spaces, but scaling to long horizons remains challenging ~\cite{moerland2020model}.
How can we combine symbolic and neural planning methods to overcome the limitations of each?

In this paper, we propose a new model-based approach for learning and planning in deterministic, goal-based, multi-task settings with continuous state and action spaces.
Following previous work, we assume that a small number of discrete \emph{predicates} (named relations over objects) are given, having been implemented by a human engineer
~\cite{lyu2019sdrl,wang2021learning}, or learned from previous experience in similar domains.
These predicates induce discrete \emph{state abstractions} of the continuous environment state.
For example, \textsc{Holding}(\texttt{block1}) abstracts away the continuous pose with which \texttt{block1} is held. Even when given predicates, the question of \emph{how to make use of them} to learn effective models for planning in continuous state and action spaces is a hard problem we seek to address.

From the predicates, and from training data of transitions in an environment, we aim to learn: (1) \emph{abstract actions}, which define transitions between abstract states; (2) an \emph{abstract transition model}, with symbolic preconditions and effects akin to AI planning operators; (3) a \emph{neural transition model} over the low-level, continuous state and action spaces; and (4) a set of \emph{neural action samplers}, which define how abstract actions can be refined into continuous actions.

We unify all of these with a new class of models that we term the \textbf{N}euro-\textbf{S}ymbolic \textbf{R}elational \textbf{T}ransition Model (NSRT) (pronounced ``insert''). NSRTs have both symbolic and neural components; all components are relational, permitting generalization to tasks with any number of objects and allowing sample-efficient learning.

To plan with NSRTs, we borrow techniques from search-then-sample task and motion planning (TAMP)~\cite{garrett2021integrated}, with symbolic AI planning in an outer loop serving as guidance for continuous planning with neural models in an inner loop. This bilevel strategy allows for fast planning in continuous state and action spaces, while avoiding the \emph{downward refinability assumption}, which would assume planning can be decomposed into separate symbolic and continuous planning steps~\cite{marthi2007angelic}.
When modeling robotic domains symbolically, the predicates are often \emph{lossy}, meaning that downward refinability cannot be assumed (\figref{fig:teaser}, top and middle).

This paper focuses on \emph{how to learn NSRTs} and \emph{how to use NSRTs for planning} in continuous-space, long-horizon tasks.
We show in four robotic planning domains, across both the PyBullet~\cite{pybullet} and AI2-THOR~\cite{ai2thor} simulators, that NSRTs are extremely data-efficient: they can be learned from a few thousand transitions. We also show that learned NSRTs allow for fast planning on new tasks, with many more objects than during training and long horizons of up to 60 actions. 
Baseline and ablation comparisons confirm that integrated neuro-symbolic reasoning is key to these successes.

%% file: related.tex
\section{Related Work}
\label{sec:related}

\textbf{Model-Based Reinforcement Learning (MBRL).}
Our work is related to MBRL in that we use data of taking actions in an environment to learn and plan with transition models.
Many recent approaches to deep MBRL learn transition models that are relatively unstructured, and therefore must resort to undirected planning strategies such as CEM~\cite{moerland2020model}.
Relational MBRL is a subfield of MBRL that uses relational learning~\cite{dvzeroski2001relational} to learn object-centric factored transition models~\cite{battaglia2016interaction} or to discover STRIPS operator models~\cite{xia2018learning,lang2012exploration} when given a set of predicates.
Our work also learns relational transition models, but with a bilevel structure that allows planning without assuming downward refinability.

\textbf{Symbolic AI Planning for RL.}
Our work continues a recent line of investigation that seeks to leverage symbolic AI planners for continuous states and actions.
For example, previous work learns propositional~\cite{dittadi2021planning,konidaris2018skills} or lifted~\cite{arora2018review,asai2019unsupervised,ahmetoglu2020deepsym} symbolic transition models, and uses them with AI planners~\cite{hoffmann2001ff,helmert2006fast}.
Other related work has used symbolic planners as managers in hierarchical RL, where low-level option policies are learned~\cite{lyu2019sdrl,illanes2020symbolic,kokelicaps21}.
This interface between symbolic planner and low-level policies assumes downward refinability, a critical assumption we do \emph{not} make.

\textbf{Learning for Hierarchical Planning.}
Reasoning at multiple levels of abstraction is a key theme in hierarchical planning~\cite{bercher2019survey}.
Task and motion planners (TAMP)~\cite{garrett2021integrated} can plan effectively at long horizons, but they typically require hand-specified operators, action samplers, and low-level transition models.
Our work continues recent research into learning these components instead~\cite{wang2021learning,zhuo2009learning,loula2020learning,silver2021learning}.
Comparatively, ours is the first to learn operators, samplers, and a low-level transition model in one unified system.

%% file: problem.tex
\section{Problem Setting}
\label{sec:problem}

We study a deterministic, goal-based, multi-task setting with continuous object-oriented states, continuous actions, and a fixed, given set of predicates.
Formally, we consider an \emph{environment}  $\langle \T, d, \A, f, \P \rangle$ and a collection of \emph{tasks}, each of which is a tuple $\langle s_0, g, H \rangle$.

\textbf{Environments.} $\T$ is a set of object types, and $d : \T \to \mathbb{N}$ defines the dimensionality of the real-valued attribute (feature) vector of each object type.
For example, an object of type \texttt{box} might have an attribute vector describing its current pose, side length, and color.
An environment state $s$ is a mapping from a set of typed objects $o$ to attribute vectors of dimension $d(o)$, where $d(o)$ is shorthand for the dimension of the attribute vector of the type of object $o$.
We use $\S$ to denote this object-oriented state space. The
$\A \subseteq \mathbb{R}^m$ is the environment action space.
The $f : \S \times \A \to \S \cup \S_{\fail}$ is a deterministic transition function mapping a state $s \in \S$ and action $a \in \A$ to either a next state in $\S$ or a failure state in $\S_{\fail}$.
A failure state ends a task attempt, and is characterized by the objects responsible for the failure, e.g., objects that are unexpectedly in collision.
\emph{Throughout this paper, the transition function $f$ is unknown to the agent.}

$\P$ is a set of \emph{predicates} given to the agent.
A predicate is a named, binary-valued relation among some number of objects. A \emph{ground atom} applies a predicate to specific objects, such as \textsc{Above}($o_1$, $o_2$), where the predicate is \textsc{Above}. A \emph{lifted atom} applies a predicate to typed placeholder variables: \textsc{Above}(?$a$, ?$b$).
Taken together, the set of ground atoms that hold in a continuous state define a \emph{discrete state abstraction}; let 
$\textsc{Abstract}(s)$ denote the abstract state for state $s \in \S$, and let $\mathcal{S}^{\uparrow}$ denote the abstract state space.
For instance, a state $s$ where objects $o_1$, $o_2$, and $o_3$ are stacked may be represented by the abstract state $\textsc{Abstract}(s) = \{\textsc{On}(o_1, o_2), \textsc{On}(o_2, o_3)\}$. Note that this abstract state loses details about the geometry of the scene.

\textbf{Tasks.} A task $\langle s_0, g, H \rangle$ is an initial state $s_0 \in \S$, a goal $g$, and a maximum horizon $H$.
We will generally denote the set of objects in $s_0$ as $\mathcal{O}$. \emph{This object set $\O$ is fixed within a task, but changes between tasks.}
Goals $g$ are sets of ground atoms over the object set $\O$, such as \{\textsc{On}($o_3$, $o_2$), \textsc{On}($o_2$, $o_1$)\}.
A solution to a task is a \emph{plan}, a sequence of at most $H$ actions $a \in \A$ such that successive application of the unknown transition model $f$, starting from $s_0$, results in a final state $s$ where $g \subseteq \textsc{Abstract}(s)$ (i.e., the goal holds).

\textbf{Data Collection and Evaluation.}
We focus on the problems of \emph{learning} and \emph{planning}.
To isolate these problems, we assume that a \emph{training dataset} of tuples $\D = \{(s, a, s')\}$ is provided, with $s \in \S$, $a \in \A$, $s' \in \S \cup \S_{\fail}$, and $f(s, a) = s'$.
The agent's objective is to maximize the number of tasks solved over a set of \emph{test tasks} that are selected to have long-horizon goals and more objects than were seen in $\D$.


%% file: representation.tex
\section{NSRT Representation}
\label{sec:representation}

The next three sections introduce Neuro-Symbolic Relational Transition Models (NSRTs).
In this section, we describe the NSRT representation; in \secref{sec:planning}, we address planning with NSRTs; and in \secref{sec:learning}, we discuss learning NSRTs.
\figref{fig:pipeline} illustrates the full pipeline.

We want models that are \emph{learnable}, \emph{plannable}, and \emph{generalizable}. To that end, we propose the following definition:

\begin{defn}
A \emph{Neuro-Symbolic Relational Transition Model (NSRT)} is a tuple $\langle \param, P, E, h, \pi \rangle$, where:
\begin{tightlist}
\item $\param = (o_1, \dots, o_k)$ is an ordered list of \emph{parameters}; each $o_i$ is a variable of some type from type set $\T$.
\item $P$ is a set of \emph{symbolic preconditions}; each precondition is a lifted atom over parameters $\param$.
\item $E = (E^+, E^-)$ is a tuple of \emph{symbolic effects}. $E^+$ are \emph{add effects}, and $E^-$ are \emph{delete effects}; both are sets of lifted atoms over parameters $\param$.
\item $h : \mathbb{R}^{d(o_1)+\dots+d(o_k)} \times \A \to \mathbb{R}^{d(o_1)+\dots+d(o_k)}$ is a \emph{low-level transition model}, a neural network that predicts next attribute values given current ones and an action.
\item $\pi(a \mid v)$ is an \emph{action sampler}, a neural network defining a conditional distribution over actions $a \in \A$, where $v~\in~\mathbb{R}^{d(o_1)+\dots+d(o_k)}$ is a vector of attribute values.
\end{tightlist}
\end{defn}

In this paper, we will learn and plan with a \emph{collection} of NSRTs. Together with the object set $\O$ of a task, a collection of NSRTs jointly defines four things: an \emph{abstract action space} for efficient planning; a (partial) \emph{abstract transition model} over the abstract state space $\S^\uparrow$ and the abstract action space; a (partial) \emph{low-level transition model} over environment states and actions; and \emph{action samplers} to refine abstract actions into environment actions.
The rest of this section describes how NSRTs define these four components.

First, we define the notion of grounding an NSRT:

\begin{defn}
Given an object set $\O$, a \emph{ground NSRT} is an NSRT whose parameters $o_i \in \param$ are replaced by objects from $\O$, following an injective \emph{substitution} $\sigma$ mapping each $o_i$ to an object.
The ground preconditions and effects under $\sigma$ are denoted $P_\sigma$ and $E_\sigma$ respectively.
\end{defn}

Given a set of NSRTs and a task with object set $\O$, the resulting set of \emph{ground} NSRTs defines an \emph{abstract} action space for that task, which we denote as $\A^\uparrow$. Therefore, the phrases \emph{abstract action} and \emph{ground NSRT} are interchangeable. For instance, say we wrote an NSRT called \textsc{Stack} with two parameters $?x$ and $?y$; let $\sigma = \{?x \mapsto o_3, ?y \mapsto o_6\}$. Then $\textsc{Stack}(o_3, o_6)$ is an abstract action with substitution $\sigma$.

Working toward a definition of the abstract transition model, we next define ground NSRT \emph{applicability}:
\begin{defn}
A ground NSRT with preconditions $P_\sigma$ is \emph{applicable} in state $s \in \mathcal{S}$ if $P_\sigma \subseteq \textsc{Abstract}(s)$.
It is also \emph{applicable} in abstract state $s^\uparrow \in \mathcal{S}^\uparrow$ if $P_\sigma \subseteq s^\uparrow$.
\end{defn}

In words, applicability simply checks that the ground NSRT's precondition atoms are a subset of the abstract state atoms.
A set of ground NSRTs defines a (partial) \emph{abstract transition model} $f^\uparrow : \mathcal{S}^\uparrow \times \mathcal{A}^\uparrow \to \mathcal{S}^\uparrow$, which maps an abstract state and abstract action (ground NSRT) to a next abstract state.
The $f^\uparrow(s^\uparrow, a^\uparrow)$ is partial since it is  only defined when $a^\uparrow$ is applicable in $s^\uparrow$; when it \emph{is} applicable, we have:
\begin{equation}\tag{Equation 1}
\label{eq:abstractmodel}
f^\uparrow(s^\uparrow, a^\uparrow) = (s^\uparrow \setminus E_\sigma^-) \cup E_\sigma^+,
\end{equation}
where $E_\sigma = (E_\sigma^+,E_\sigma^-)$ are the effects for $a^\uparrow$.
In words, this abstract transition model removes delete effects and includes add effects, as long as the preconditions of the ground NSRT are satisfied.
This symbolic representation is akin to operators in classical AI planning~\cite{bonet2001planning}; we use this connection to our advantage in \secref{sec:planning}.

What is the connection between the symbolic components of an NSRT ($P$ and $E$) and the environment transitions? To answer this question, we use the following definition:

\begin{defn}
\label{defn:covered}
A ground NSRT $a^\uparrow$ with effects $(E_\sigma^+, E_\sigma^-)$ \emph{covers} an environment transition $\tau = (s, a, s')$, denoted $a^\uparrow \models \tau$, if (1) the ground NSRT is applicable in $s$; (2) $E_\sigma^+ = \textsc{Abstract}(s')\setminus\textsc{Abstract}(s)$; and (3) $E_\sigma^- =  \textsc{Abstract}(s)\setminus\textsc{Abstract}(s')$.
\end{defn}

We assume that the following \emph{weak semantics} connect $P$ and $E$ with the environment: for each ground NSRT $a^\uparrow$, there exists a state $s \in \mathcal{S}$ and there exists an action $a \in \mathcal{A}$ such that $a^\uparrow \models (s, a, f(s, a))$. Importantly, this means that the abstraction defined by the NSRTs does \emph{not} satisfy downward refinability~\cite{marthi2007angelic}, which would have required the ``there exists a state'' to be ``for all states.'' These weak semantics make learning efficient (\secref{sec:learning}), but require integrated planning (\secref{sec:planning}).

To plan, it is important to be able to simulate the effects of actions on the continuous environment state. 
We next discuss the low-level transition model $h$, which is used for this.

\begin{defn}
\label{defn:context}
Given a state $s$ and ground NSRT $a^\uparrow$ with substitution $\sigma$, the \emph{context of $s$ for $a^\uparrow$} is $v_\sigma(s) = s[\sigma(o_1)] \circ \cdots \circ s[\sigma(o_k)]$, where $v_\sigma(s) \in \mathbb{R}^{d(o_1)+\dots+d(o_k)}$, $s[\cdot]$ looks up an object's attribute vector in $s$, and $\circ$ is vector concatenation.
\end{defn}

In words, the context for a ground NSRT is the subset of a state's attribute vectors that correspond to the ground NSRT's objects, assembled into a vector.
The context is the input to the low-level neural transition model $h$:
$$h(v_\sigma(s), a) \approx v_\sigma(f(s, a)),$$
where, recall, $f$ is the \emph{unknown} environment transition model.
All objects not in $\sigma$ are predicted to be unchanged.

Finally, the neural action sampler $\pi$ of an NSRT connects the abstract and environment action spaces: it samples continuous actions from the environment action space $\A$ that lead to the NSRT's symbolic effects.
Given a state $s$ and applicable ground NSRT with substitution $\sigma$, if $a \sim \pi(\cdot \mid v_\sigma(s))$, then $(s, a, f(s, a))$ should ideally be covered by the ground NSRT.
The fact that $\pi$ is stochastic can be useful for planning, where multiple samples may be required to achieve desired effects (see Figure~\ref{fig:teaser}, or \cite{wang2021learning}).

\emph{There are three key properties of NSRTs to take away from these definitions.}
(1) NSRTs are fully relational, i.e., invariant over object identities.
This leads to data-efficient learning and generalization to novel tasks and objects.
(2) NSRTs do not assume downward refinability, as discussed above.
(3) NSRTs are \emph{locally scoped}; all components of a ground NSRT are defined only where it is applicable.
This modularity leads to independent learning problems; see Section~\ref{sec:learning}.

%% file: planning.tex
\section{Neuro-Symbolic Planning with NSRTs}
\label{sec:planning}

\begin{figure*}[t]
  \centering
  \noindent
    \includegraphics[width=\textwidth]{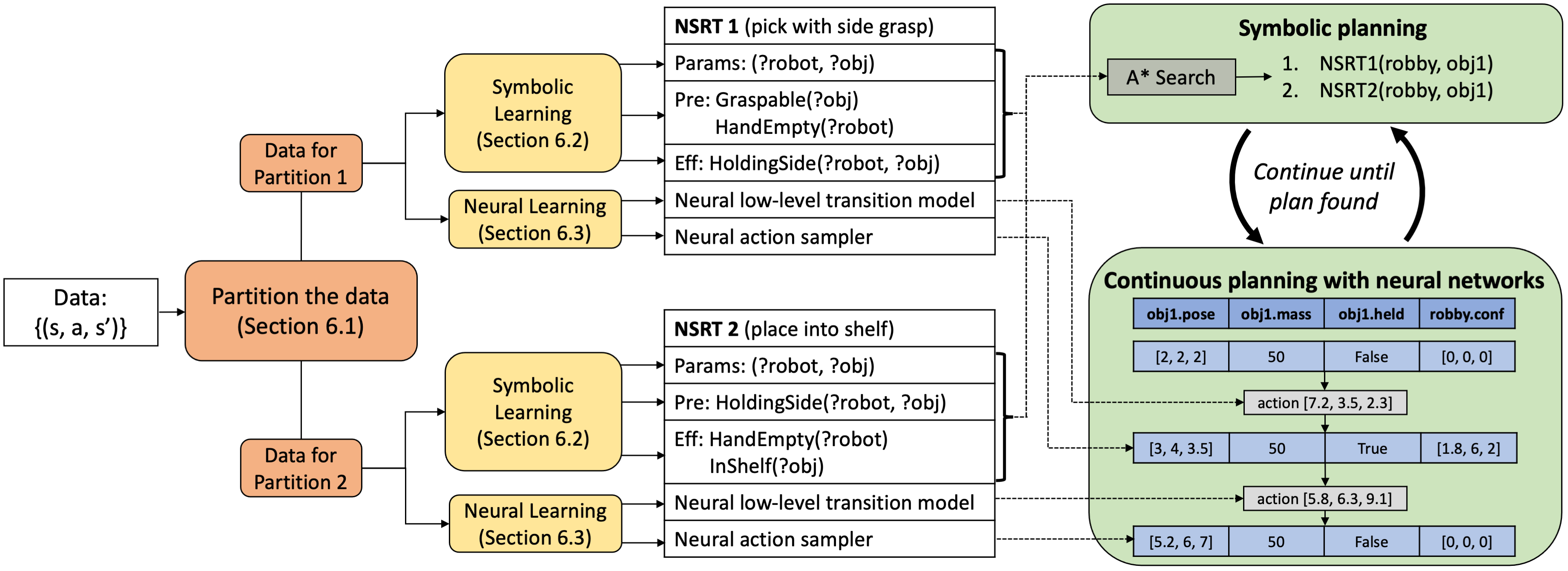}
    \caption{\small Our pipeline, with a simplified Painting example. An NSRT (\secref{sec:representation}) contains both symbolic components used for A$^*$ search with AI planning heuristics, and neural components used for continuous planning. The example NSRTs shown in the middle require that a robot must be side-grasping an object to place it into a shelf. These NSRTs are \emph{not} ground: their parameters are variables, so these NSRTs can be applied to any objects. We learn NSRTs from transition data (\secref{sec:learning}), and then use them to perform bilevel planning (\secref{sec:planning}). Delete effects are omitted from this figure for visual clarity.}
  \label{fig:pipeline}
\end{figure*}

\input{pseudocode/planning_pseudocode}

We now describe how NSRTs can be used to plan in a given task.
Recall that the weak semantics of NSRTs (\secref{sec:representation}) do \emph{not} guarantee downward refinability: a sequence of abstract actions that achieves a goal cannot necessarily be turned into a sequence of environment actions achieving that goal.
Our strategy will be to perform integrated bilevel planning, with an outer search in the abstract space informing an inner loop producing environment actions. This planning strategy falls under the broad class of search-then-sample TAMP techniques~\cite{garrett2021integrated}.
\emph{See \algref{alg:planning} for pseudocode.}

\textbf{Symbolic Planning.} We perform an outer A$^*$ search from $\textsc{Abstract}(s_0)$ to $g$, with the abstract transition model of \ref{eq:abstractmodel} and uniform action costs. For the search heuristic, we use $h_{\text{add}}$, a domain-independent heuristic from the symbolic planning literature~\cite{bonet2001planning} that approximates the state-to-goal distance under a delete relaxation of the abstract model.
This A$^*$ search will find candidate \emph{symbolic plans}: sequences of ground NSRTs $a^\uparrow \in \A^\uparrow$.

\textbf{Continuous Planning.} For each candidate symbolic plan, an inner loop attempts to refine it into a \emph{plan} --- a sequence of actions $a \in \mathcal{A}$ that achieves the goal $g$ --- using the neural components of the NSRTs.
We use the action sampler $\pi$ and low-level transition model $h$ of each ground NSRT in the symbolic plan to construct an \emph{imagined} state-action trajectory starting from the initial state $s_0$.
If the goal $g$ holds in the final imagined state, we are done. If $g$ does not hold, or if any state's abstraction does not equal the expected abstract state according to the A$^*$ search, then we repeat this process. After $n_\text{trials}$ (a hyperparameter) unsuccessful imagined trajectories, we return control to the A$^*$ search.

\textbf{Handling Failures.} Recall that some transitions in the environment can lead to a failure state in $\S_{\fail}$.
Following \cite{tampinterface}, we would like to use the presence of a failure state during continuous planning to inform symbolic planning. 
We begin by assuming that we have a model which predicts whether a failure state is reached, and if so, the set of objects $\{o_1, \ldots, o_j\}$ that were responsible for the failure (e.g., two objects that were in collision, or an object that broke irreparably); we will show how to learn this model in \secref{sec:learning}. Now, we perform a domain-independent procedure: 
we introduce special predicates \textsc{NotCausesFailure} for every object type in the environment, and for each NSRT, we add a symbolic effect $\textsc{NotCausesFailure}(o_i)$ for each $o_i$ in the parameters $\param$. This says that every action affecting a set of objects absolves all those objects from being responsible for a failure; we found this simple technique sufficient for our experiments, but other, more domain-specific information can be leveraged instead~\cite{tampinterface}. During refinement, if a failure is predicted, we terminate the inner loop, update the preconditions of the ground NSRT at that timestep to include $\{\textsc{NotCausesFailure}(o_1), \ldots, \textsc{NotCausesFailure}(o_j)\}$ (where $\{ o_1, \ldots, o_j \}$ are the set of objects predicted to be responsible for the failure), and restart A$^*$ from the initial state. This forces A$^*$ to either consider actions which change the states of these objects before using the same ground NSRT, or just avoid using this ground NSRT entirely.

%% file: pseudocode/planning_pseudocode.tex
\begin{algorithm}[t]
  \SetAlgoNoEnd
  \DontPrintSemicolon
  \SetKwFunction{algo}{algo}\SetKwFunction{proc}{proc}
  \SetKwProg{myalg}{Algorithm}{}{}
  \SetKwProg{myproc}{Subroutine}{}{}
  \SetKw{Continue}{continue}
  \SetKw{Break}{break}
  \SetKw{Return}{return}
  \myalg{\textsc{Bilevel Planning with NSRTs}}{
    \nonl \textbf{Input:} NSRT set $\{\langle O, P, E, h, \pi \rangle\}$\;
    \nonl \textbf{Input:} Task $\langle s_0, g, H \rangle$\;
    \nonl \textbf{Input:} $n_{\text{trials}}$: \# of imagined trajectory tries\;
    \tcp{\footnotesize A$^*$ with symbolic components of NSRTs and classical heuristics.}
    \nonl $s_0^\uparrow \gets \textsc{Abstract}(s_0)$\;
    \nonl \For{$\overline{p} \in \text{A}^*(s_0^\uparrow, g, H, \{\langle O, P, E, \cdot, \cdot \rangle\})$}{
    \nonl \For{$n_{\text{trials}}$ tries}{
    \nonl Initialize \texttt{plan} as empty list\;
    \tcp{\footnotesize Imagine rollout with neural components of ground NSRTs.}
    \nonl $s \gets s_0$\;
    \nonl \For{ground NSRT $\langle \cdot, \cdot, \cdot, \pi, h \rangle \in \overline{p}$}{
    \nonl $a \sim \pi(\cdot \mid s)$ \tcp{\footnotesize stochastic}
    \nonl Append $a$ to \texttt{plan}\;
    \nonl $s \gets h(s, a)$\;
    }
    \nonl \If{$g \subseteq \textsc{Abstract}(s)$}{
    \nonl \Return \texttt{plan}}
    }}
    }\;
\caption{Pseudocode for bilevel planning with NSRTs. The outer loop runs A$^*$ search over the symbolic components of the NSRTs, from the symbolic initial state $s_0^\uparrow = \textsc{Abstract}(s_0)$ to the goal $g$. This A$^*$ produces candidate symbolic plans $\overline{p}$, which are sequences of ground NSRTs. The neural components of these ground NSRTs are used in the inner loop, which tries $n_\text{trials}$ times to \emph{refine} a symbolic plan into a sequence of continuous actions from the environment action space $\A$. If the goal $g$ holds in the final state, we are done. In practice, we perform an extra optimization (not shown): we terminate the inner loop early whenever \textsc{Abstract}($s$) deviates from the expected sequence of states under $\overline{p}$.
}
\label{alg:planning}
\end{algorithm}

%% file: learning.tex
\section{Learning NSRTs}
\label{sec:learning}

We now address the problem of learning the structure (\secref{subsec:partitioning}), the symbolic components (\secref{subsec:learningsymbolic}), and the neural components (\secref{subsec:learningneural}) of NSRTs, all using the training dataset $\D$ (\secref{sec:problem}).

\subsection{Partitioning the Transition Data}
\label{subsec:partitioning}

Recall that $\D$ contains a set of samples from the unknown transition model $f$: each sample is a state $s \in \S$, an action $a \in \A$, and either a
next state in $s' \in \S$ or a failure state in $\S_{\fail}$. We will ignore the transitions that led to $\S_{\fail}$; they will be used in \secref{subsec:learnfailure}.
We begin by partitioning the set of transitions $\tau = (s, a, s')$ so that each partition $\psi \in \Psi$ will correspond to a single NSRT, thus automatically determining the number of learned NSRTs.
Two transitions belong to the same partition iff their symbolic effects can be \emph{unified}:
\begin{defn}
Two transitions $\tau_1$ and $\tau_2$ can be \emph{unified} if there exists a bijective mapping $\sigma$ from the objects in $\textsc{Eff}(\tau_1)$ to the objects in $\textsc{Eff}(\tau_2)$ s.t. $\sigma[\textsc{Eff}(\tau_1)] = \textsc{Eff}(\tau_2)$, where $\textsc{Eff}(\tau) = (\textsc{Abstract}(s') \setminus \textsc{Abstract}(s), \textsc{Abstract}(s) \setminus \textsc{Abstract}(s'))$, and $\sigma[\cdot]$ denotes substitution following $\sigma$.
\end{defn}

These partitions can be computed in time linear in the number of transitions, objects, and atoms per effect set.

\subsection{Learning the Symbolic Components}
\label{subsec:learningsymbolic}

We now show how to learn NSRT parameters $\param$, symbolic preconditions $P$, and symbolic effects $E$ for each partition $\psi \in \Psi$.
First, we define a mapping \textsc{Ref} that maps a transition $\tau$ to a subset of objects in $\tau$ that are ``involved'' in the transition.
In practice, we implement $\textsc{Ref}(\tau)$ by selecting all objects that appear in $\textsc{Eff}(\tau)$.\footnote{This suffices for our experiments, but it cannot capture ``indirect effects,'' where some objects influence a transition without themselves changing; other implementations of \textsc{Ref} could be used instead.}
By construction of our partitions, every transition $\tau \in \psi$ will have equivalent $\textsc{Ref}(\tau)$, up to object renaming.
We thus introduce NSRT parameters $\param$ corresponding to the types of all the objects in any arbitrarily chosen transition's $\textsc{Ref}(\tau)$. For each $\tau \in \psi$, let $\sigma_\tau$ be a bijective mapping from these parameters $\param$ to the objects in $\textsc{Ref}(\tau)$.
The NSRT symbolic effects follow by construction: $E = \sigma_\tau^{-1}[\textsc{Eff}(\tau)]$ for any arbitrary $\tau \in \psi$.

To learn the symbolic preconditions $P$ for the NSRT corresponding to partition $\psi$, we use a simple inductive approach that restricts learning by assuming that for each lifted effect set seen in the data, there is exactly one lifted precondition set.\footnote{See \cite{silver2021learning} for a more expensive method that avoids this assumption.}
By this assumption, the preconditions follow from an intersection of projected abstract states: $$P = \bigcap_{\tau = (s, \cdot, \cdot) \in \psi} \sigma_{\tau}^{-1}[\textsc{Project}(\textsc{Abstract}(s))],$$
where \textsc{Project} maps $\textsc{Abstract}(s)$ to the subset of atoms whose objects are all in $\textsc{Ref}(\tau)$.
By construction, the semantics we defined in \secref{sec:representation} are satisfied over the training dataset: each transition belongs to one partition, and the preconditions for that partition must hold in its abstract state.

\subsection{Learning the Neural Components}
\label{subsec:learningneural}
We now describe how to learn a low-level transition model $h$ and action sampler $\pi$ for each partition's NSRT.
The key idea is to use the state projections computed during partitioning to create regression problems.
Recalling Definition~\ref{defn:context}, let $v_\sigma = s[\sigma_\tau(o_1)] \circ \dots \circ s[\sigma_\tau(o_k)]$
denote the context of state $s$ from transition $\tau$, where $(o_1, o_2, \ldots, o_k)$ are the NSRT parameters.
In words, $v_{\sigma}$ is a vector of the attribute values in state $s$ corresponding to the objects that map the ground atoms $\textsc{Eff}(\tau)$ of the transition to the lifted effects $E$ of the NSRT.
We can do the same to produce $v_{\sigma'}$ for $s'$. 
Applying this to all transitions in $\psi$ gives us a dataset of $(v_\sigma, a, v_{\sigma'})$.

Recall that we want to learn $h$ such that $h(v_\sigma(s), a) \approx v_\sigma(f(s, a))$.
With the dataset above, this learning problem now reduces to regression, with $v_\sigma$ and $a$ being the inputs and $v_{\sigma'}$ being the output. We use a fully connected neural network (FCN) as the regressor, trained to minimize mean-squared error.
Learning $\pi$ requires \emph{distribution} regression, where we  fit $P(a \mid v_\sigma)$ to the transitions $(v_\sigma, a, \cdot)$.
We use an FCN that takes $v_\sigma$ as input and predicts the mean $\mu$ and covariance matrix $\Sigma$ of a Gaussian.
This FCN is trained to maximize the likelihood of action $a$ under $\N(\mu, \Sigma)$.\footnote{Here, we are assuming that the desired action distribution has nonzero measure. In practice, $\Sigma$ can be arbitrarily small.}
Since Gaussians have limited expressivity, we also learn an \emph{applicability classifier} that maps pairs $(v_\sigma, a)$ to 0 or 1, implemented as an FCN with binary cross-entropy loss. We implement $\pi$ as a rejection sampler that draws from the Gaussian until the applicability classifier returns a 1.\footnote{If the applicability classifier fails enough times (10 in experiments), we terminate the inner loop and continue the outer A$^*$ search (see \algrefshort{alg:planning}).}

\subsection{Learning to Predict Failures}
\label{subsec:learnfailure}
Here we address the problem of learning to anticipate failures during planning.
Note that unlike NSRT learning (\secref{sec:learning}), which is ``locally scoped'' to a fixed number of objects defined by the NSRT parameters, failure prediction can require reasoning about all objects in the full state.
Extracting the training data transitions that led to a failure state in $\S_{\fail}$, we create a dataset of the form $\{(s, a, \ofail)\}$, where $\ofail$ is the set of objects in the failure state.
We then train a graph neural network (GNN) that takes as input $s$, $\textsc{Abstract}(s)$, and $a$, and outputs a score between 0 and 1 for each object, representing the predicted probability that it is included in $\ofail$.
We follow \cite{ploi} for the graph encoding and GNN architecture.
Once trained, we use the GNN to predict both whether a transition to a failure state occurs and $\ofail$, by checking whether there are any objects whose score is over 0.5, and including them in $\ofail$ if so.

%% file: experiments.tex
\section{Experiments}
\label{sec:experiments}

Our empirical evaluations address the following key questions:
\textbf{(Q1)} Can NSRTs be learned data-efficiently?
\textbf{(Q2)} Can learned NSRTs be used to plan to long horizons, especially in tasks involving new and more objects than were seen in the training dataset?
\textbf{(Q3)} Is bilevel planning efficient and effective, and are both levels needed?
\textbf{(Q4)} To what extent are learned action samplers useful for planning?

\subsection{Experimental Setup}
\label{subsec:expsetup}
We evaluate Q1-Q4 by running seven methods on four environments. All experiments were run on Ubuntu 18.04 using 4 CPU cores of an Intel Xeon Platinum 8260 processor.

\textbf{Environments.} In this section, we describe our four environments.
The environments are illustrated in \figref{fig:teaser} (bottom row).
Each environment has two sets of tasks: ``easy'' test and ``hard'' test. ``Hard'' test tasks require generalization to more objects. In all environments, we transition to a failure state in $\S_\fail$ whenever a geometric collision occurs.
\begin{tightlist}
\item \emph{Environment 1:} In ``PickPlace1D,'' a robot must pick blocks and place them into designated target regions on a table.
All poses are 1D. Some placements are obstructed by movable objects; none of the predicates capture obstructions, causing a lack of downward refinability.
\item \emph{Environment 2:} In ``Kitchen,'' a robot waiter in 3D must pick cups, fill them with water, wine, or coffee, and serve them to customers.
Some cups are too heavy to be lifted; the cup masses are not represented by the predicates, causing a lack of downward refinability.
\item \emph{Environment 3:} In ``Blocks,'' a robot in 3D must stack blocks on a table to make towers.
In this environment only, the downward refinability assumption holds.
\item \emph{Environment 4:} In ``Painting,'' a robot in 3D must pick, wash, dry, paint, and place widgets into a box or shelf. Placing into the box (resp. shelf) requires picking with a top (resp. side) grasp. All widgets must be painted a particular color before being placed, which first requires washing/drying if the widget starts off dirty or wet. The box has a lid that may obstruct placements; whether the lid will obstruct a placement is not represented symbolically, causing a lack of downward refinability.
\end{tightlist}

\begin{figure*}[t]
  \centering
  \noindent
    \includegraphics[width=\textwidth]{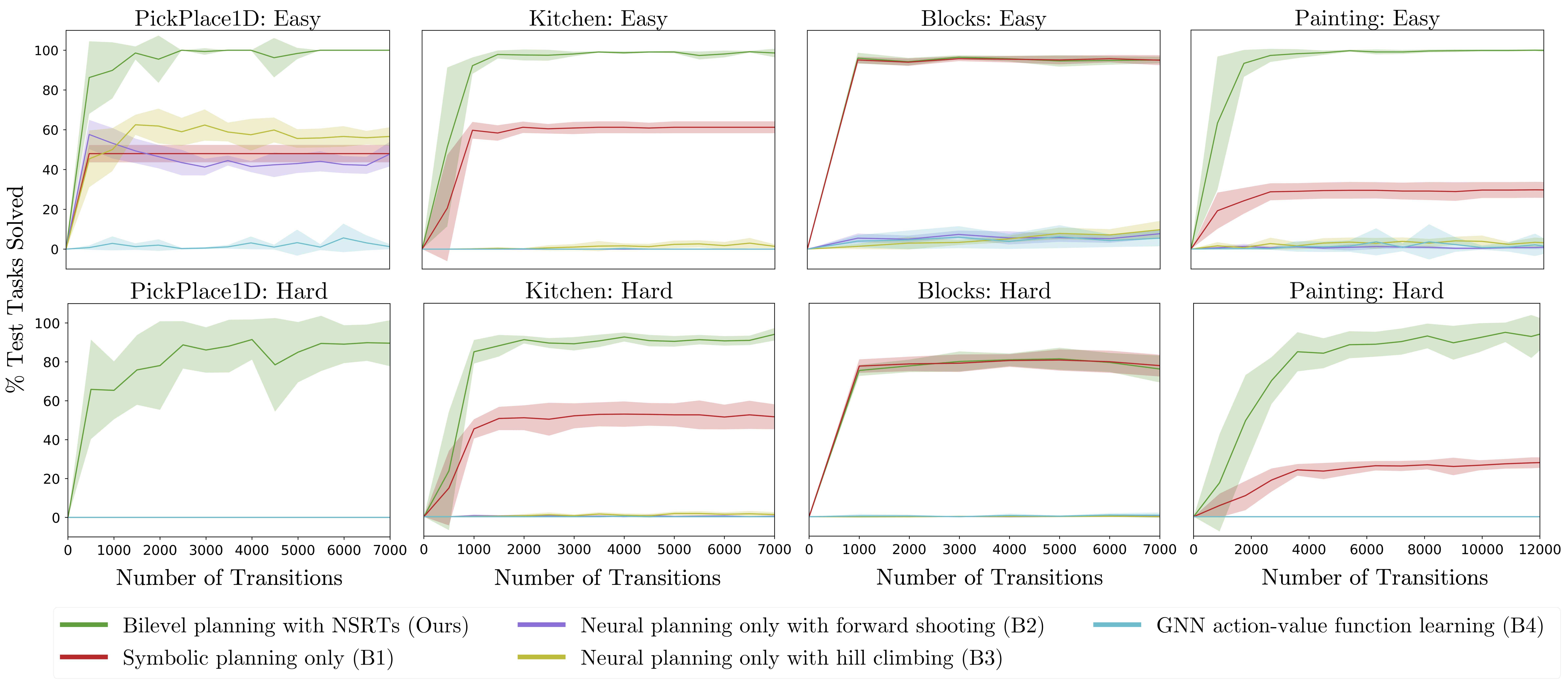}
    \caption{\small Learning curves showing the percentage of 100 randomly generated test tasks (top row: easy tasks; bottom row: hard tasks) solved versus the number of transitions in the dataset. Each curve depicts a mean over 8 seeds, with standard deviation shaded. All methods have a timeout of 3 seconds per task. NSRTs (green) quickly learn to solve many more tasks than the baselines, especially in the hard tasks.}
  \label{fig:mainresults}
\end{figure*}

\textbf{Dataset Creation.}
We create the training dataset $\D$ for each environment by, 700 times, sampling an initial state $s_0$ and running a scripted stochastic policy $\pi_0$ from it.
This policy avoids the zero-measure issues that would arise from uniformly random actions, but it is not goal-directed, and it does not nearly suffice to solve test tasks. For PickPlace1D, Kitchen, and Blocks, we experiment with up to 7000 transitions; for Painting, up to 12000 since it is more challenging.

\textbf{Methods Evaluated.} We evaluate the following methods. \emph{Note that B4-B6 receive information that Ours and B1-B3 do not have access to: the scripted stochastic policy $\pi_0$ mentioned above.} All methods get the same training dataset.

\begin{tightlist}
    \item \emph{Ours: Bilevel planning with NSRTs.} This is our main approach. Plans are executed open-loop.
    \item \emph{B1: Symbolic planning only.} This baseline performs symbolic planning using the symbolic components of the learned NSRTs. When a symbolic plan is found that reaches the goal, it is immediately executed by calling the learned action samplers for the corresponding ground NSRTs in sequence, open-loop. The low-level transition models are not used. This baseline ablates away our integrated planner and assumes downward refinability.
    \item \emph{B2: Neural planning only with forward shooting.} This baseline randomly samples $H$-length sequences of ground NSRTs and uses their neural components to imagine a trajectory, repeating until it finds a trajectory where the final state satisfies the goal. This baseline does not use the symbolic components of the NSRTs, and thus can be seen as an ablation of the symbolic planning.
    \item \emph{B3: Neural planning only with hill climbing.} This baseline performs local search over full plans. At each iteration, a random plan step is resampled using the learned action sampler of a random NSRT. The new plan is rejected unless it improves the number of goal atoms satisfied in the final imagined state.
    As in B2, the symbolic components of the NSRTs are not used.
    \item \emph{B4: GNN action-value function learning.} This ``model-free'' baseline trains a goal-conditioned graph neural network (GNN) action-value function using fitted Q-iteration. The GNN takes as input a continuous low-level state, the corresponding abstract state, and a continuous action; it outputs a value. At evaluation time, given a state, we draw candidate actions from $\pi_0$ (see above).
    \item \emph{B5: No learned samplers.} This baseline is an ablation of our main approach that does not use the learned NSRT action samplers $\pi$. Instead, actions are drawn from $\pi_0$.
    \item \emph{B6: No symbolic components or learned samplers.} This baseline is an ablation that uses the forward shooting of B2 but with actions drawn from $\pi_0$, like B5. Only the low-level transition models $h$ are used.
\end{tightlist}

\textbf{Additional details.} All neural networks are fully connected with two hidden layers of size 32, and trained using the Adam optimizer for 35K (action samplers), 10K (low-level transition models), or 50K (applicability classifier) epochs with a learning rate of 1e-3.
In our robotic environments of interest, transitions are often \emph{sparse}, changing only a subset of object attributes at any given time.
For learning the low-level transition model, we exploit this by calculating the attributes that change in \emph{any} transition within a partition, and only predict next values for those attributes.
For learning the action samplers, we restrict the covariance matrix $\Sigma$ to be diagonal and positive semi-definite using an exponential linear unit~\cite{clevert2015fast}.
During evaluation, we clip samples from the action samplers to be at most 1 standard deviation from the mean, for improved stability.
The applicability classifier is trained with negative examples collected from either data in other partitions, or data in the same partition but with the objects re-mapped.
We subsample negative examples to ensure that the dataset is balanced in a 1:1 ratio with the positive examples.
In all experiments, we use $n_\text{trials}=1$, which we found to be sufficient due to the accuracy of the action samplers and low-level transition models.
For the action-value function (B4), we train by running 5 iterations of fitted Q-iteration, and during evaluation, we sample 100 candidate actions from the scripted policy $\pi_0$ at each step, choosing the action with the best predicted value to execute in the environment.
Methods that use shooting (B2 and B6) try up to 1000 iterations, or until the timeout (3 seconds for every method across all experiments) is reached.

\input{ablation_table}

\subsection{Results and Discussion}

See \figref{fig:mainresults} for learning curves.
The main observation is that in all environments, our method quickly learns to solve tasks within the allotted 3-second timeout. Thus, \textbf{Q1} and \textbf{Q2} can be answered affirmatively. Turning to \textbf{Q3}, we can study whether bilevel planning is effective by comparing Ours, B1, and B2.
The gap between Ours and B1 shows the importance of integrated bilevel planning. B1 will not be effective in any environment where downward refinability does not hold; only Blocks is downward refinable, which explains the identical performance of Ours and B1 there. B2 fails in most cases, confirming the usefulness of the symbolic components of the learned NSRTs.

Both B3 and B4 are generally ineffective. B3 performs local search, which is much weaker than our directed A$^*$. B4 is model-free, forgoing planning in favor of learning a value function; such strategies are known to be more data-hungry~\cite{moerland2020model}.
In our experimentation, we found that for the Easy test tasks, B4 starts performing decently after seeing about four times as much data as we used in making the plots, confirming that it requires substantially more data.

To evaluate \textbf{Q4}, we turn to an ablation study. \tabref{tab:ablations} compares our method with B5 and B6, both of which sample actions from the scripted policy $\pi_0$ rather than using our learned NSRT action samplers. First, comparing B5 and B6, bilevel planning is much better than shooting, which speaks to the benefits of using the symbolic components of the NSRTs to guide planning; this conclusion was also supported by \figref{fig:mainresults}. Second, comparing Ours and B5, the learned action samplers help substantially. This is because $\pi_0$ is highly generic, not targeted toward any specific set of effects like our NSRT action samplers are.

%% file: ablation_table.tex
\begin{table*}[t]
	\centering
	\footnotesize
	\begin{tabular}{| l | p{0.68cm} | p{0.68cm} | p{0.68cm} | p{0.68cm} | p{0.68cm} | p{0.68cm} | p{0.68cm} | p{0.68cm} | }
	\hline
	\multicolumn{1}{|c|}{} &\multicolumn{2}{c|}{PickPlace1D} &
	\multicolumn{2}{c|}{Kitchen} &
	\multicolumn{2}{c|}{Blocks} &
	\multicolumn{2}{c|}{Painting} \\
	\hline
	\emph{Methods} &
	{\scriptsize Easy} & {\scriptsize Hard} &
	{\scriptsize Easy} & {\scriptsize Hard} &
	{\scriptsize Easy} & {\scriptsize Hard} &
	{\scriptsize Easy} & {\scriptsize Hard} \\
	\hline
    Bilevel planning with NSRTs (Ours) & \textbf{98.4} & \textbf{85.0} & \textbf{99.1} & \textbf{90.4} & \textbf{95.0} & \textbf{79.6} & \textbf{99.6} & \textbf{89.6} \\
    No learned samplers (B5) & \textbf{95.9} & 46.4 & 71.9 & 32.6 & 89.9 & 53.4 & 84.5 & 0.1 \\
    No symbolic components or learned samplers (B6) & 71.1 & 0.0 & 0.0 & 1.5 & 62.9 & 8.6 & 5.4 & 0.0 \\
    \hline
	\end{tabular}
	 \caption{\small Percentage of 100 randomly generated test tasks solved after learning on the full training dataset. Each number is a mean over 8 seeds; bold results are within one standard deviation of best. Both the symbolic components \emph{and} the learned samplers are critical.}
	\label{tab:ablations}
\end{table*}

%% file: conclusion.tex
\section{Conclusion, Limitations, and Future Work}
\label{sec:conclusion}

We proposed NSRTs for long-horizon, goal-based, object-oriented planning tasks, showed that their neuro-symbolic structure affords fast bilevel planning, and found that they are data-efficient to learn and effective at generalization, outperforming several baselines.
Key limitations of this work include the assumption that predicates are given and the assumption that environments are deterministic and fully observable. To address the former, NSRTs could be combined with work on learning predicates from high-dimensional inputs \cite{asai2019unsupervised}.
For the latter, we hope to draw on TAMP techniques for stochastic and partially observed settings. 